\documentclass[letterpaper]{article} 
\usepackage{aaai2026}  
\usepackage{times}  
\usepackage{helvet}  
\usepackage{courier}  
\usepackage[hyphens]{url}  
\usepackage{graphicx} 
\usepackage{natbib}  
\usepackage{caption} 
\usepackage{algorithm}
\usepackage{algorithmic}
\usepackage{subcaption}
\usepackage{amssymb}
\usepackage{amsmath}
\usepackage{booktabs}
\usepackage{multirow}
\usepackage{pifont}
\usepackage{times}
\usepackage{helvet}
\usepackage{courier}
\usepackage{xcolor}
\usepackage{newfloat}
\usepackage{listings}
\DeclareCaptionStyle{ruled}{labelfont=normalfont,labelsep=colon,strut=off} 
\floatstyle{ruled}
\newfloat{listing}{tb}{lst}{}
\floatname{listing}{Listing}
\title{M3SR: Multi-Scale Multi-Perceptual Mamba for Efficient Spectral Reconstruction}
\author{
    Yuze Zhang\textsuperscript{\rm 1}, Lingjie Li\textsuperscript{\rm 2}\thanks{Corresponding author.}, Qiuzhen Lin\textsuperscript{\rm 1}\footnotemark[1], Zhong Ming\textsuperscript{\rm 1}, Fei Yu\textsuperscript{\rm 1}, Victor C. M. Leung\textsuperscript{\rm 1}\\
}
\affiliations{
    \textsuperscript{\rm 1}College of Computer Science and Software Engineering, Shenzhen University, Shenzhen, China\\
    \textsuperscript{\rm 2}School of Artificial Intelligence, Shenzhen Technology University, Shenzhen, China\\

    \{zhangyuze2021, qiuzhlin, mingz, yufei, vleung\}@szu.edu.cn, lilingjie@sztu.edu.cn
}

\usepackage{bibentry}

\begin{document}

\maketitle

\begin{abstract}
	The Mamba architecture has been widely applied to various low-level vision tasks due to its exceptional adaptability and strong performance. 
	Although the Mamba architecture has been adopted for spectral reconstruction, it still faces the following two challenges: (1) Single spatial perception limits the ability to fully understand and analyze hyperspectral images; (2) Single-scale feature extraction struggles to capture the complex structures and fine details present in hyperspectral images.
	To address these issues, we propose a multi-scale, multi-perceptual Mamba architecture for the spectral reconstruction task, called M3SR. 
	Specifically, we design a multi-perceptual fusion block to enhance the ability of the model to comprehensively understand and analyze the input features. 
	By integrating the multi-perceptual fusion block into a U-Net structure, M3SR can effectively extract and fuse global, intermediate, and local features, thereby enabling accurate reconstruction of hyperspectral images at multiple scales.
	Extensive quantitative and qualitative experiments demonstrate that the proposed M3SR outperforms existing state-of-the-art methods while incurring a lower computational cost.
\end{abstract}

\begin{links}
    \link{Code}{https://github.com/zhangyuzecn/M3SR}
\end{links}

\section{Introduction}
Hyperspectral imaging uses narrow wavelength bands to gather detailed spatial and spectral data, providing a more comprehensive view of scenes compared to RGB images. 
This enhanced detail renders it valuable across various domains, including environmental monitoring~\cite{Lassalle2021,Wang2024}, medical imaging~\cite{Leon2023,Liu2024}, and agriculture~\cite{Khan2022,Ram2024}. 
Capturing hyperspectral images (HSIs) is a very time-consuming and complex process, but fortunately Spectral Reconstruction (SR) provides an efficient solution by generating HSIs from RGB images. 
Earlier SR methods, based on traditional techniques such as sparse dictionaries~\cite{Arad2016}, Gaussian processes~\cite{Akhtar2018}, and low-rank representations~\cite{Gao2020}, struggle to recognize complex patterns or adapt to different scenarios. 
In contrast, the advent of deep learning, particularly convolutional neural networks (CNNs), has greatly improved SR performance, although challenges remain in handling long-range connections and ensuring consistent spectral detail. 
Moreover, the Transformer with its multi-head self-attention mechanism effectively tackles these issues by capturing distant relationships and fine spectral variations, making it a powerful tool for SR. 
However, the computational requirements of the Transformer increase sharply with image size, posing efficiency challenges for HSIs with both detailed spatial resolution and broad spectral resolution.

\begin{figure}[t]
	\centering
	\includegraphics[scale=0.38]{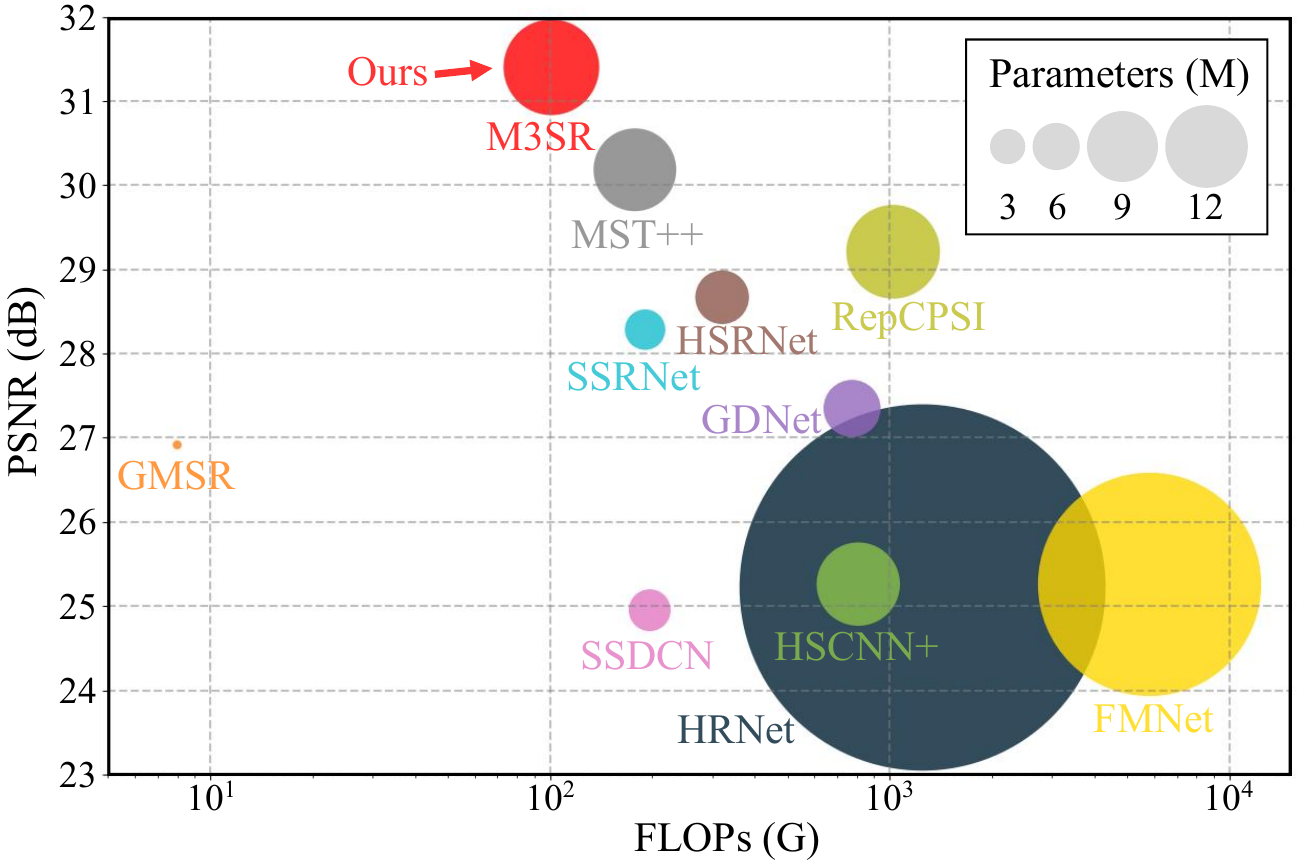}
	\caption{PSNR \emph{v.s.} Parameters \emph{v.s.} FLOPs comparison with existing spectral reconstruction methods. For an intuitive analysis, FLOPs and PSNR are represented by the horizontal and vertical axis, and the circle radius indicates parameters. The proposed M3SR achieves a better balance between PSNR and parameters and FLOPs.}
	\label{fig:Bubble}
\end{figure}

Recently, the novel architecture Mamba~\cite{Gu2023}, based on the state space model (SSM), has quickly risen in popularity within natural language processing. 
Compared to the Transformer, Mamba exhibits superior efficiency in processing extended sequences, while consistently delivering robust performance during both the training and inference phases~\cite{Patro2024}. 
Inspired by this success, VMamba~\cite{Liu2024a} integrates SSM-based blocks into visual networks to enhance the learning of visual features. 
As a result, Mamba-based methods have gained wide acceptance for tasks such as semantic segmentation~\cite{Wan2024}, object detection~\cite{Wang2024b}, and HSI classification~\cite{Li2024}. 
While Mamba-based methods are effective in modelling long-range dependencies, they are fundamentally based on processing pixel sequences and have limitations in terms of feature perception capabilities. 
Therefore, the application of the Mamba architecture to SR tasks still faces the following two challenges: (1) Single spatial perception limits the ability to fully understand and analyze HSIs; (2) Single-scale feature extraction struggles to capture the complex structures and detailed information in HSIs.

To address these challenges, we design a Multi-scale Multi-perceptual Mamba architecture for SR, called M3SR. 
Specifically, we introduce a Multi-Perceptual Fusion (MPF) block to enhance the ability of the model to comprehensively understand and analyze the input features. 
Furthermore, by integrating the MPF block into the U-Net structure~\cite{Ronneberger2015}, M3SR is able to capture complex structures and detailed information in HSIs. 
The architecture of M3SR integrates multi-scale and multi-perceptual feature extraction capabilities to facilitate efficient SR by capturing and fusing both detailed and global features across spatial, frequency, and spectral dimensions at multiple scales. 
As shown in Fig.~\ref{fig:Bubble}, the proposed M3SR strikes a suitable balance between reconstruction accuracy and computational cost. 
Finally, our M3SR model achieves superior performance compared to state-of-the-art (SOTA) super-resolution models on four public benchmark datasets, while using fewer parameters and requiring less computational resources. The main contributions of this paper are as follows.

\begin{itemize}
	\item A multi-perceptual fusion (MPF) block is designed, which adaptively integrates spatial, frequency, and spectral information, significantly enhancing the richness and robustness of feature representation.
	\item A multi-scale, multi-perceptual Mamba architecture (M3SR) is proposed, which is capable of extracting perceptual features at global, intermediate, and local scales, and effectively fusing multi-scale information for efficient SR.
	\item Extensive quantitative and qualitative experiments demonstrate that our proposed M3SR achieves superior performance compared to state-of-the-art (SOTA) methods, with significantly fewer parameters and lower computational cost (FLOPs).
\end{itemize}

\section{Related Work}
\subsection{Spectral Reconstruction from RGB Images}
SR technology aims to reconstruct HSIs from RGB images, tackling challenges such as high acquisition costs and low spatial resolution. 
Existing SR methods can be categorized into traditional methods, CNN-based methods, and Transformer-based methods. 
Traditional methods use statistical priors for SR, such as sparse dictionaries~\cite{Arad2016}, Gaussian processes~\cite{Akhtar2018}, and low-rank representations~\cite{Gao2020}, but they are limited in reconstructing high-quality HSI~\cite{Wu2023}. 
CNN-based methods use CNNs to map RGB to HSI, with models such as HSCNN+~\cite{Shi2018} and HRNet~\cite{Zhao2020} delivering strong results, although they struggle to capture long-range dependencies. 
Transformer-based methods (e.g., MST++~\cite{Cai2022} and ESSAformer~\cite{Zhang2023}) excel in long-range correlation modelling, but the high computational cost limits their application. 
Therefore, we design M3SR based on Mamba, which can capture long-range dependencies with linear complexity to achieve efficient SR.

\subsection{State Space Models for Low-level Vision}
Recently, SSMs are increasingly applied to low-level vision tasks, with remarkable results~\cite{Zhang2024,Wu2024,Weng2025}. 
MambaIR~\cite{Guo2025} first introduces SSMs to image restoration and provides a simple baseline for related research. 
Subsequently, MambaFormerSR~\cite{Zhi2024} combines SSMs with the Transformer architecture and demonstrates superior performance in super-resolution tasks for remote sensing images. 
SRMamba-T~\cite{Liu2025} further innovates by designing a multi-directional selective scanning module and a feature fusion module, achieving an excellent balance between computational efficiency and performance in single-image super-resolution tasks.

However, these methods represent only the initial exploration of SSMs in low-level vision tasks. 
Their single-scale feature extraction mechanisms struggle to fully capture the complex structures and detailed information in images.  
To address this, U-net is integrated with SSMs due to its multi-scale feature extraction capabilities, which leads to significant advances in several low-level vision tasks. 
For example, in dehazing~\cite{Zheng2024}, deblurring~\cite{Liu2024c, Kong2024}, and image restoration~\cite{Deng2024a, Sepehri2024}, the combination of SSMs and U-net shows excellent performance. 
Meanwhile, thanks to the advantages of frequency domain analysis in localization and multi-resolution capabilities, the integration of SSMs with frequency domain analysis also shows excellent results in several low-level vision tasks. For example, the combination of SSMs and frequency domain analysis achieves remarkable success in super-resolution~\cite{Xiao2024}, deraining \cite{Li2024a} and magnetic resonance imaging reconstruction \cite{Zou2024a}.

Nevertheless, the task of reconstructing HSIs from RGB images using SSMs still faces numerous challenges. 
GMSR~\cite{Wang2024a} is the first method to combine SSMs with spatial and spectral gradients for SR tasks. While GMSR significantly improves computational efficiency, it relies solely on SSMs to extract spatial information, limiting the ability of SSMs to comprehensively understand and analyze images. 
Additionally, its single-scale feature extraction mechanism struggles to balance global semantic information with local details. 
Therefore, we propose a multi-scale, multi-perceptual Mamba architecture for SR tasks, aiming to enable comprehensive understanding and analysis of images, as well as to effectively capture complex structures and fine-grained details.

\section{Preliminaries}
\begin{figure}[t]
	\centering
	\begin{minipage}{0.4\linewidth}
		\centering
		\begin{subfigure}{\linewidth}
			\includegraphics[width=\linewidth]{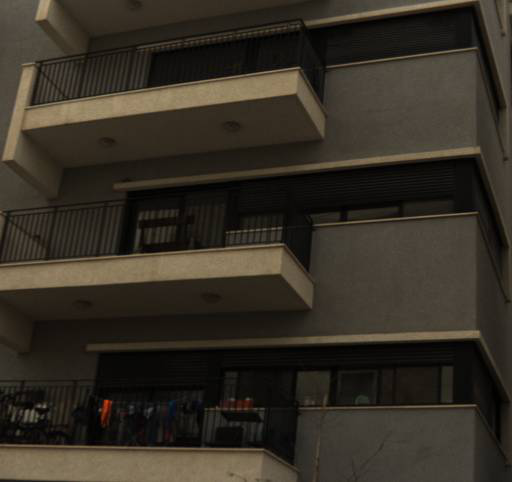}
			\caption{Input Image.}
			\label{fig:DWT iuput}
		\end{subfigure}
	\end{minipage}
	\hfill
	\begin{minipage}{0.55\linewidth}
		\centering
		\begin{subfigure}{0.45\linewidth}
			\includegraphics[width=\linewidth]{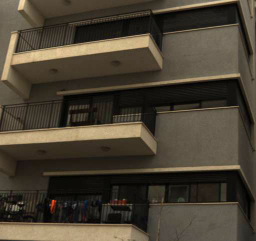}
			\caption{LL.}
			\label{fig:LL}
		\end{subfigure}
		\hfill
		\begin{subfigure}{0.45\linewidth}
			\includegraphics[width=\linewidth]{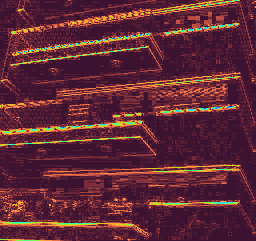}
			\caption{LH.}
			\label{fig:LH}
		\end{subfigure}
		\vspace{0.5cm} 
		\begin{subfigure}{0.45\linewidth}
			\includegraphics[width=\linewidth]{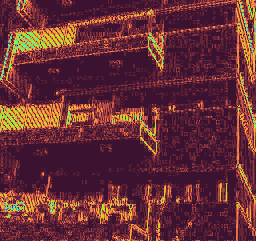}
			\caption{HL.}
			\label{fig:HL}
		\end{subfigure}
		\hfill
		\begin{subfigure}{0.45\linewidth}
			\includegraphics[width=\linewidth]{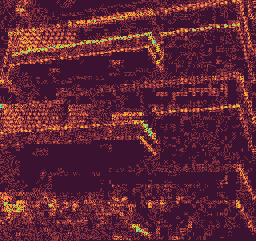}
			\caption{HH.}
			\label{fig:HH}
		\end{subfigure}
	\end{minipage}
	\caption{Input image and sub-bands after discrete wavelet transform decomposition.}
	\label{fig:DWT}
\end{figure}

\begin{figure*}[t]
	\centering
	\begin{subfigure}{0.47\linewidth}
		\includegraphics[width=\linewidth]{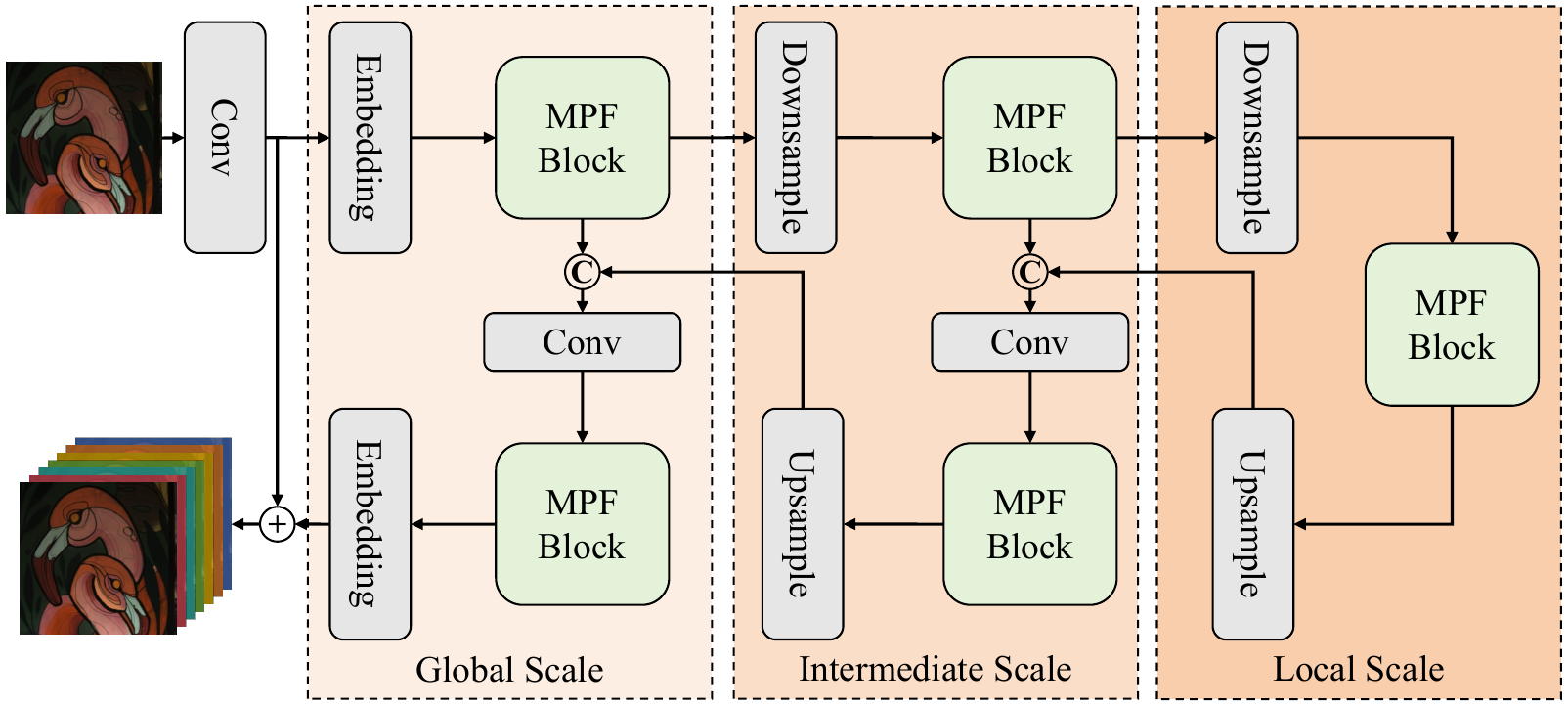}
		\vspace{-1mm}
		\caption{The architecture of our M3SR.}
		\label{fig:Architecture}
	\end{subfigure}
	\begin{subfigure}{0.45\linewidth}
		\hspace{7mm}
		\includegraphics[width=\linewidth]{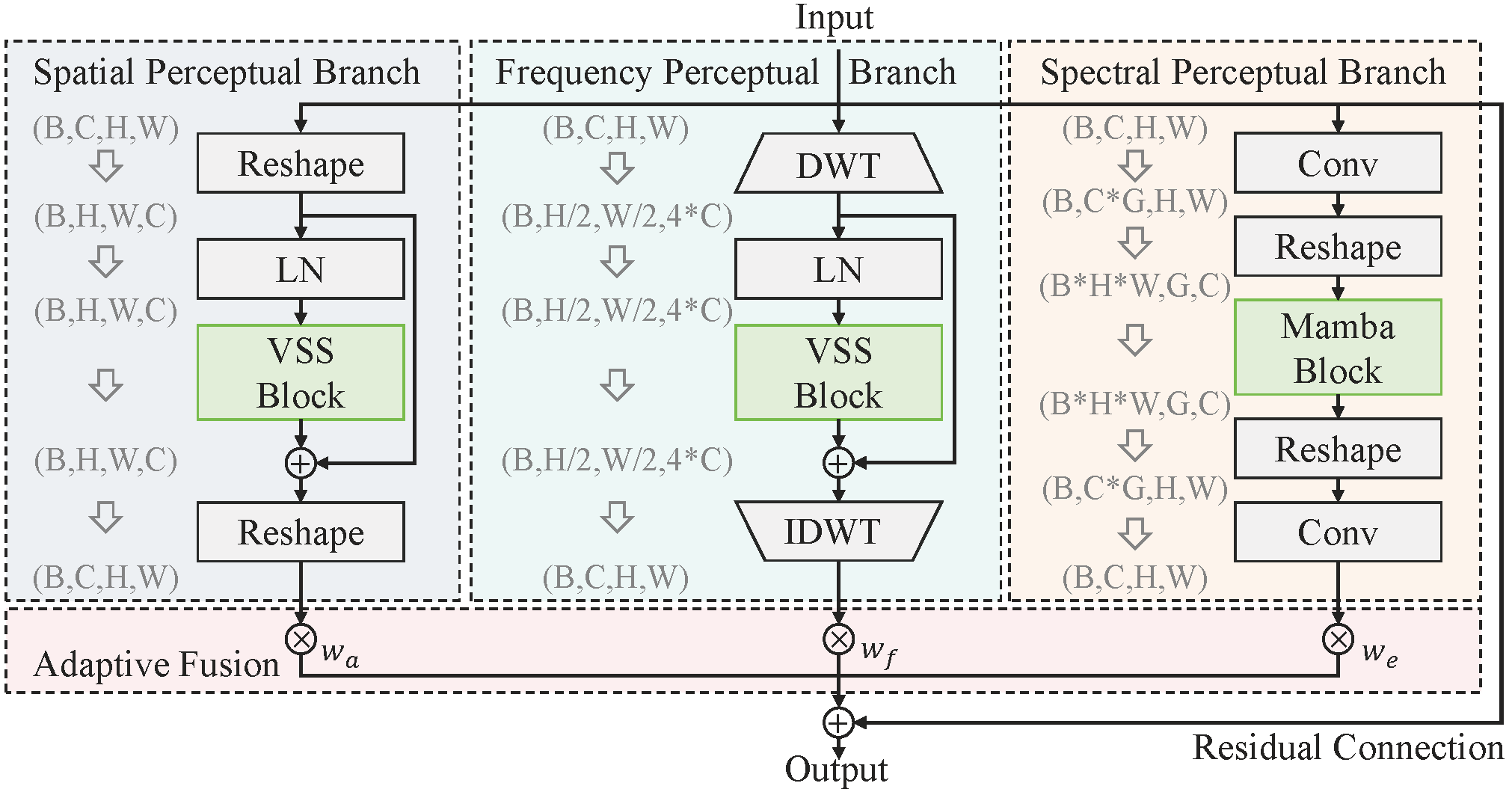}
		\caption{The structure of the MPF Block.}
		\label{fig:MPF Block}
	\end{subfigure}
	\caption{The architecture of our M3SR and the structure of the multi-perceptual fusion (MPF) block.}
	\label{fig:M3SR}
\end{figure*}

\textbf{Discrete Wavelet Transform:} The DWT is a highly effective method of image processing, allowing detailed analysis in both the time and frequency domains to accurately capture localized image features~\cite{Latif2024}. 
By applying the Inverse Discrete Wavelet Transform (IDWT) to the transformed signal, the original image can be accurately reconstructed using the wavelet coefficients. 
The DWT decomposes an input image $I \in \mathbb{R}^{H \times W \times C}$ into four distinct sub-bands: $I_{LL}$, $I_{LH}$, $I_{HL}$, and $I_{HH}$. This decomposition can be expressed as: 
\begin{equation}
	I_{LL}, I_{LH}, I_{HL}, I_{HH} = DWT(I),
\end{equation}
where $I_{LL}$ represents a low-frequency approximation suitable for image scaling, while $I_{LH}$, $I_{HL}$, and $I_{HH}$ emphasize high-frequency details such as edges in different orientations. 
As shown in Fig.~\ref{fig:DWT}, after DWT decomposition, the resulting sub-bands contain texture information at different resolutions. 
By performing analysis and processing of these sub-bands, key texture features in the image can be effectively extracted, leading to more efficient and accurate SR.

\textbf{State Space Model: }SSM is a framework designed to describe the state representation of a sequence at each time step and predict its next state based on given inputs. It operates by mapping a one-dimensional input function or sequence \( x(t) \in \mathbb{R} \) to an output \( y(t) \in \mathbb{R} \) through a hidden state representation \( h(t) \in \mathbb{R}^N \). This transformation can be mathematically formulated using a linear ordinary differential equation as shown below:

\begin{align}
	h'(t) &= \mathbf{A} h(t) + \mathbf{B} x(t), \\
	y(t) &= \mathbf{C} h(t).
\end{align}
where \( \mathbf{A} \in \mathbb{R}^{N \times N} \) represents the state matrix, \( \mathbf{B} \in \mathbb{R}^{N \times 1} \) and \( \mathbf{C} \in \mathbb{R}^{1 \times N} \) denote the projection parameters.

To make the continuous system compatible with deep learning, the zero-order hold (ZOH) technique is employed, using a time-scale parameter \( \Delta \) to facilitate discretization. This approach simplifies the conversion of the continuous parameters \( \mathbf{A} \) and \( \mathbf{B} \) into their discrete counterparts \( \overline{\mathbf{A}} \) and \( \overline{\mathbf{B}} \). After discretization, the discretized SSM system can be formulated as follows:

\begin{align}
	h_t &= \overline{\mathbf{A}} h_{t-1} + \overline{\mathbf{B}} x_t, \\
	y_t &= \mathbf{C} h_t. 
\end{align}

To enhance the computational efficiency and scalability, the convolution operation is harnessed to expedite the linear recurrence process outlined above. Consequently, the ultimate output can be synthesized as

\begin{align}
	\overline{\mathbf{K}} &= \big(\mathbf{C} \overline{\mathbf{B}}, \mathbf{C} \overline{\mathbf{A}} \overline{\mathbf{B}}, \ldots, \mathbf{C} \overline{\mathbf{A}}^{L-1} \overline{\mathbf{B}} \big), \\
	y &= x * \overline{\mathbf{K}},
\end{align}
where \( L \) is the length of the input sequence \( x \), and \( \overline{\mathbf{K}} \in \mathbb{R}^L \) represents a structured convolutional kernel.

To tackle the limitation of SSMs in capturing contextual information, Mamba introduces a selection mechanism and proposes the selective SSM (S6) to achieve dynamic interactions between sequential states. 
Additionally, Mamba introduces a hardware-aware algorithm that computes the model recurrently with a scan, ensuring both effectiveness and efficiency in capturing global contextual information.

\section{Methodology}

\subsection{The Overall Architecture}
The architecture of M3SR is illustrated in Fig.~\ref{fig:Architecture}. 
Based on the U-net structure, M3SR uses an encoder-decoder structure to efficiently extract and fuse multi-scale features. 
The architecture progressively extracts multi-scale semantic features by downsampling operations and restores spatial resolution by upsampling operations, forming a symmetric feature processing workflow. 
The multi-scale feature extraction mechanism of M3SR is divided into three scales: global scale, intermediate scale and local scale. 
The global scale captures the overall structure, the intermediate scale focuses on contextual information of the HSI, while the local scale concentrates on recovering detailed texture information. 

The MPF block is designed as the fundamental block of M3SR. 
As shown in Fig.~\ref{fig:MPF Block}, the MPF block achieves feature extraction across multiple perceptual domains through three parallel branches. 
The first branch focuses on spatial perception, capturing the spatial structure and geometric features. The second branch uses DWT for frequency perception, extracting multiple frequency features. 
The third branch targets spectral perception, enhancing the ability of the MPF block to model spectral features. 
Through this multi-perceptual feature extraction and fusion mechanism, M3SR efficiently integrates spatial, frequency and spectral feature information. 
It maintains the overall structural consistency of the image while accurately recovering detailed texture and high-frequency information, ultimately achieving high-quality reconstruction of HSIs.

\begin{figure}[t]
	\centering
	\begin{subfigure}{0.4\linewidth}
		\hspace{7mm}
		\includegraphics[width=0.55\linewidth]{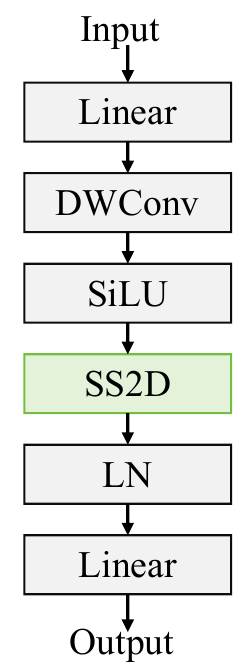}
		\caption{VSS Block.}
		\label{fig:VSS Block}
	\end{subfigure}
	\begin{subfigure}{0.4\linewidth}
		\includegraphics[width=\linewidth]{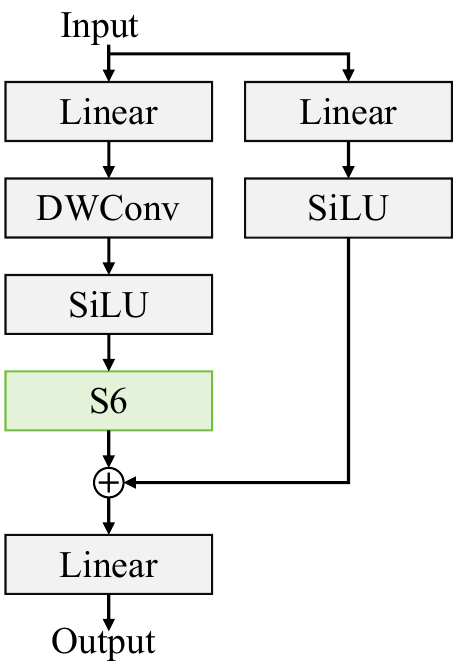}
		\caption{Mamba Block.}
		\label{fig:Mamba Block}
	\end{subfigure}
	\caption{The structures of VSS block and Mamba block.}
	\label{fig:short}
\end{figure}

\subsection{Multi-perceptual   Fusion Block}
As a core component, the MPF block is proposed to overcome the limitations of single perception. 
By exploiting the perception of multi-dimensional information, including spatial, frequency, and spectral domains, more accurate and comprehensive SR is achieved. 
As shown in Fig.~\ref{fig:MPF Block}, there are three main branches, spatial perceptual, frequency perceptual, and spectral perceptual branches, which are introduced as follows.

\textbf{Spatial Perceptual Branch:} 
In this branch, we employ the Mamba architecture as the foundational component to develop a spatial perceptual branch capable of modeling long-range dependencies with linear computational efficiency. 
Although the core mechanism of Mamba and its parallel selective scanning process is highly effective, it is primarily designed for one-dimensional sequential data processing. 
Therefore, the 2D Selective-Scan (SS2D) technique from VMamba is used to capture two-dimensional spatial information. 
Specifically, the SS2D technique transforms a 2D image into a 1D sequence and conducts scans along four distinct directions to extract long-range dependencies within each sequence. 
These sequences are subsequently aggregated through summation and reconfigured into the original 2D format. 
The structure of the VSS block is presented in Fig.~\ref{fig:VSS Block}, formulated as:

\begin{align}
	VSS(x) &= Lin(LN(SS2D(x'))),\\
	x' &= SiLU(DWConv(Lin(x))),
\end{align}
where $x$ is the input, $Lin$ is the linear layer, $SiLU$ is the SiLU activation function, $DWConv$ is the depth-wise convolution and $SS2D$ is the 2D Selective-Scan module.

Subsequently, the spatial perceptual branch is designed based on the VSS block to improve the spatial feature extraction. 
The input feature $\mathbf{F}_{in}$ is first reshaped to obtain the expanded feature $\mathbf{F}^1_a$. 
This feature is then processed by the VSS block, which extracts spatial information from the HSI. 
Finally, the features are reshaped to generate the output feature $\mathbf{F}^2_a$, as defined below:

\begin{align}
	\mathbf{F}^1_a &= Reshape(\mathbf{F}_{in}),\\
	\mathbf{F}^2_a &= Reshape(Concat(VSS(LN(\mathbf{F}^1_a)), \mathbf{F}^1_a)),
\end{align}
where $LN$ is the layer norm, and $VSS$ is the VSS block. 

\textbf{Frequency Perceptual Branch:} 
To comprehensively and accurately extract texture features from HSIs, a frequency perceptual branch is designed to fully leverage information from different frequency domains. 
Specifically, DWT is used to decompose the input feature $\mathbf{F}_{in}$ into sub-bands, resulting in one low-frequency component and three high-frequency components to separate the local details from the overall information. 
Then, the VSS block is used to perform feature learning on these frequency components to capture global contextual information from the low-frequency component while extracting local detail features from the high-frequency components. 
Finally, Inverse Discrete Wavelet Transform (IDWT) is used to reconstruct the processed frequency components into the original representation $\mathbf{F}^2_a$, as defined below:

\begin{align}
	\mathbf{F}^1_f &= DWT(\mathbf{F}_{in}),\\
	\mathbf{F}^2_f &= IDWT(Concat(VSS(LN(\mathbf{F}^1_f)),\mathbf{F}^1_f)),
\end{align}
where $DWT$ is the Discrete Wavelet Transform, and $IDWT$ is the Inverse Discrete Wavelet Transform. 


\textbf{Spectral Perceptual Branch: }
Unlike traditional vision systems that rely on RGB channels for image capture, HSIs offer a wider spectral range and finer spectral resolution. 
The spectral data associated with each pixel can be interpreted as a continuous sequence with approximate continuity across wavelengths. 
Consequently, the spectral perceptual branch is built on the basis of the Mamba block to effectively model the dependencies within the spectral data and extract distinctive features that improve the analysis. 
The structure of the Mamba block, illustrated in Fig.~\ref{fig:Mamba Block}, is mathematically defined as:
\begin{align}
	Mamba(x) &= Lin(x' + x''),\\
	x' &= SiLU(Lin(x)),\\
	x'' &= S6(SiLU(DWConv(Lin(x)))),
\end{align}
where $x$ is input and $S6$ is the selective SSM (S6).

Using this Mamba block, the spectral perceptual branch can capture relationships across the spectral dimension. 
For an input feature $\mathbf{F}_{in}$, the channel dimension $C$ is first expanded to $C \times G$ by a convolutional operation, and the features are segmented into $G$ groups by reshaping. 
The Mamba block is then used to characterize the interactions and obtain $\mathbf{F}_e^1$, refining the spectral information in the process. 
The refined spectral feature is processed by convolution and reshaping operations to obtain $\mathbf{F}_e^2$, as defined below:

\begin{align}
	\mathbf{F}^1_e &= Mamba(Reshape(Conv(\mathbf{F}_{in}))),\\
	\mathbf{F}^2_e &= Conv(Reshape(\mathbf{F}^1_e)),
\end{align}
where $Mamba$ is the Mamba block.

\textbf{Adaptive Fusion:} 
Finally, the multi-perceptual features extracted from the spatial perceptual, frequency perceptual, and spectral perceptual branches are adaptively fused to achieve an accurate reconstruction of the HSI. 
For this purpose, three adaptive weights are used to evaluate the importance of the spatial, frequency, and spectral perceptual features and fuse them with a weighted manner. 
In addition, we incorporate a residual connection to mitigate overfitting that may occur during training. 
The fusion process is expressed as follows:

\begin{equation}
	\begin{aligned}
		\mathbf{F}_{out} &= \omega_a * \mathbf{F}^2_a + \omega_f * \mathbf{F}^2_f + \omega_e * \mathbf{F}^2_e + \mathbf{F}_{in},
	\end{aligned}
\end{equation}
where $\omega_a$, $\omega_f$, and $\omega_e$ denote the fusion weights of spatial, frequency, and spectral features, respectively. Note that $\omega_a$, $\omega_f$, and $\omega_e$ are randomly initialized and updated through back propagation to determine the final fusion weights.

\subsection{Loss Function}

The M3SR network optimization employs the mean absolute error (MAE) as the loss function, striking a balance between simplicity and effectiveness. 
Despite not using a more complex loss strategy, the network shows strong convergence during training, demonstrating its ability to capture fine image detail while maintaining structural integrity. 
Let the reconstructed HSI be \( Z \in \mathbb{R}^{H \times W \times C} \) and the reference HSI be \( \hat{Z} \in \mathbb{R}^{H \times W \times C} \). The loss function \( L \) is defined as:

\begin{equation}
	L = \frac{1}{H \times W \times C} \sum_{i=1}^{H} \sum_{j=1}^{W} \sum_{k=1}^{C} \left| Z_{i,j,k} - \hat{Z}_{i,j,k} \right|,
\end{equation}
where \( H \), \( W \), and \( C \) denote the height, width, and spectral channels of the image, respectively. 


\section{Experiments}

\begin{table*}[t]
	\centering
	\resizebox{\textwidth}{!}{
		\begin{tabular}{lcccccccccc}
			\toprule
			\multirow{2}{*}{Methods} & \multirow{2}{*}{Params (M)} & \multirow{2}{*}{FLOPs (G)} & \multicolumn{4}{c}{NTIRE2022} & \multicolumn{4}{c}{CAVE} \\ 
			\cmidrule(lr){4-7} \cmidrule(lr){8-11}
			& & & RMSE($\downarrow$) & PSNR($\uparrow$) & SAM($\downarrow$) & MSSIM($\uparrow$) & RMSE($\downarrow$) & PSNR($\uparrow$) & SAM($\downarrow$) & MSSIM($\uparrow$) \\ 
			\midrule
			HSCNN+~\cite[CVPR]{Shi2018}      & 1.642  & 807.998   & 0.058  & 25.2622  & 5.9151  & 0.8576   & 0.0227  & 33.8117  & 6.9895  & 0.9718   \\
			FMNet~\cite[AAAI]{Zhang2020}     & 11.793 & 5819.522  & 0.0586 & 25.2593  & 6.8735  & 0.863    & 0.0247  & 33.1439  & 8.5393  & 0.9706   \\
			HRNet~\cite[CVPR]{Zhao2020}      & 31.705 & 1249.005  & 0.0577 & 25.2223  & \textbf{\underline{6.055}}   & 0.8772   & \textbf{0.0205}  & 34.531   & \textbf{6.6962}  & 0.9787   \\
			GDNet~\cite[TCI]{Zhu2021}       & 0.772  & 774.456   & 0.0455 & 27.3483  & 8.1031  & 0.8841   & 0.0291  & 31.3349  & 8.6479  & 0.9685   \\
			HSRNet~\cite[TNNLS]{He2021}       & 0.683  & 321.078   & 0.0438 & 28.6688  & 6.4454  & 0.9111   & 0.0227  & 33.3828  & 7.8339  & 0.9775   \\
			SSDCN~\cite[TGRSL]{Chen2021}      & 0.415  & 196.525   & 0.0623 & 24.9525  & 7.9265  & 0.8501   & 0.0218  & 34.2542  & 7.1912  & 0.9778   \\
			MST++~\cite[CVPR]{Cai2022a}      & 1.62   & 177.737   & \textbf{0.035}  & \textbf{30.1822}  & 6.0892  & \textbf{0.9308}   & \textbf{0.0205}  & \textbf{34.6534}  & \textbf{\underline{6.6725}}  & \textbf{\underline{0.9826}}   \\
			RepCPSI~\cite[TGRS]{Wu2023a}     & 2.078  & 1023.235  & 0.0409 & 29.2083  & 6.4858  & 0.9181   & 0.0229  & 33.6155  & 7.4338  & 0.9772   \\
			SSRNet~\cite[TNNLS]{Dian2023}     & \textbf{0.388}  & 190.479   & 0.0465 & 28.2843  & \textbf{6.1205}  & 0.9118   & 0.0252  & 32.7655  & 7.4848  & 0.9767   \\
			GMSR~\cite[arXiv]{Wang2024a}      &\textbf{\underline{0.019}}  & \textbf{\underline{7.988}}     & 0.0492 & 26.9156  & 8.2357  & 0.8798   & 0.0206  & 34.5797  & 7.418   & 0.9769   \\
			M3SR [Ours]                      & 2.166  & \textbf{100.924}  & \textbf{\underline{0.0343}} & \textbf{\underline{31.3995}}  & 6.6199  & \textbf{\underline{0.9351}}   & \textbf{\underline{0.0184}}  & \textbf{\underline{35.6063}}  & 6.7887  & \textbf{0.9814}   \\
			\bottomrule
	\end{tabular}}
	\caption{Reconstruction results on the NTIRE2022 and CAVE datasets. The \textbf{\underline{champion}} and \textbf{second place} for each metric are marked in underlined bold and bold, respectively.}
	\label{tab:NTIRE2022}
\end{table*}

\begin{table*}[t]
	\centering
	\resizebox{\textwidth}{!}{
		\begin{tabular}{lcccccccccc}
			\toprule
			\multirow{2}{*}{Methods} & \multirow{2}{*}{Params (M)} & \multirow{2}{*}{FLOPs (G)} & \multicolumn{4}{c}{NTIRE2020-Clean} & \multicolumn{4}{c}{NTIRE2020-realworld} \\ 
			\cmidrule(lr){4-7} \cmidrule(lr){8-11}
			& & & RMSE($\downarrow$) & PSNR($\uparrow$) & SAM($\downarrow$) & MSSIM($\uparrow$) & RMSE($\downarrow$) & PSNR($\uparrow$) & SAM($\downarrow$) & MSSIM($\uparrow$) \\ 
			\midrule
			HSCNN+~\cite[CVPR]{Shi2018}       & 1.642  & 807.998   & 0.0274  & 32.7659  & 3.628  & 0.9555   & 0.0281  & 31.5774  & 4.3754  & 0.9427   \\
			FMNet~\cite[AAAI]{Zhang2020}      & 11.793 & 5819.522  & 0.0241  & 33.634  & 3.6259  & 0.9757   & 0.0247  & 32.8718  & 4.2207  & 0.9583   \\
			HRNet~\cite[CVPR]{Zhao2020}       & 31.705 & 1249.005  & 0.0309  & 32.0203  & \textbf{3.5785}  & 0.9617   & 0.0282  & 32.1507  & \textbf{4.1579}  & 0.9527   \\
			GDNet~\cite[TCI]{Zhu2021}         & 0.772  & 774.456   & \textbf{\underline{0.0196}}  & 36.1221  & 3.9366  & 0.9777   & 0.0192  & 35.3121  & 4.687  & 0.9611   \\
			HSRNet~\cite[TNNLS]{He2021}       & 0.683  & 321.078   & \textbf{0.0198}  & 37.1743  & 3.6409  & \textbf{\underline{0.9841}}   & 0.0213  & 34.5468  & 4.6183  & 0.9592   \\
			SSDCN~\cite[TGRSL]{Chen2021}      & 0.415  & 196.525   & 0.0216  & 35.7702  & 3.9704  & 0.9777   & 0.0199  & 35.0647  & 4.3352  & 0.96   \\
			MST++~\cite[CVPR]{Cai2022a}       & 1.62  & 177.737    & \textbf{0.0198}	 &\textbf{36.324}	 &3.3886   &0.9833   & \textbf{0.0185}  & \textbf{35.6348}  & 4.3617  & \textbf{0.9636}   \\
			RepCPSI~\cite[TGRS]{Wu2023a}      & 2.078  & 1023.235  & 0.0212  & 36.2077  & 3.6616  & 0.9802   & 0.0203  & 34.9522  & \textbf{\underline{4.0163}}  & 0.9625   \\
			SSRNet~\cite[TNNLS]{Dian2023}     & \textbf{0.388}  & 190.479   & 0.0239  & 35.0742  & 3.6473  & 0.9757   & 0.021  & 35.0337  & 4.2956  & 0.9591   \\
			GMSR~\cite[arXiv]{Wang2024a}      & \textbf{\underline{0.019}}  & \textbf{\underline{7.988}}   & 0.0239  & 33.9653  & 3.9759  & 0.9644    & 0.0278  & 31.9018  & 4.5457  & 0.9456   \\
			M3SR [Ours]                       & 2.166  & \textbf{100.924}   & \textbf{\underline{0.0196}}  & \textbf{\underline{37.7088}}  & \textbf{\underline{3.2842}}  & \textbf{\underline{0.9841}}   & \textbf{\underline{0.0171}}  &\textbf{\underline{36.3524}}  &4.2843  &\textbf{\underline{0.964}}   \\
			\bottomrule
	\end{tabular}}
	
	\caption{Reconstruction results on the NTIRE2020-Clean and NTIRE2020-Realworld datasets. The \textbf{\underline{champion}} and \textbf{second place} for each metric are marked in underlined bold and bold, respectively.}
	\label{tab:NTIRE2020}
\end{table*}

\begin{figure*}[t]
	\centering
	\begin{subfigure}[b]{0.14\textwidth}
		\includegraphics[width=\textwidth]{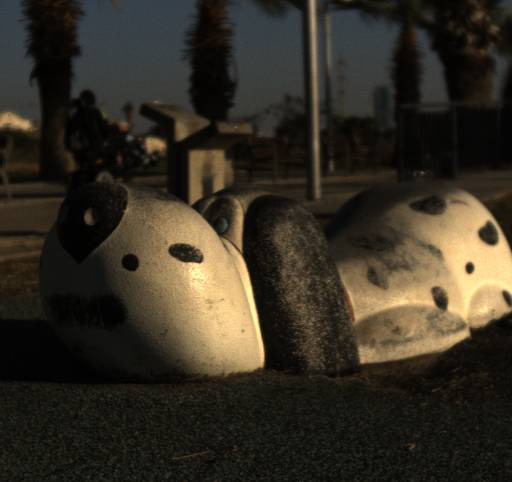}
		\caption{Input}
		\label{fig:input}
	\end{subfigure}
	\hfill
	\begin{subfigure}[b]{0.14\textwidth}
		\includegraphics[width=\textwidth]{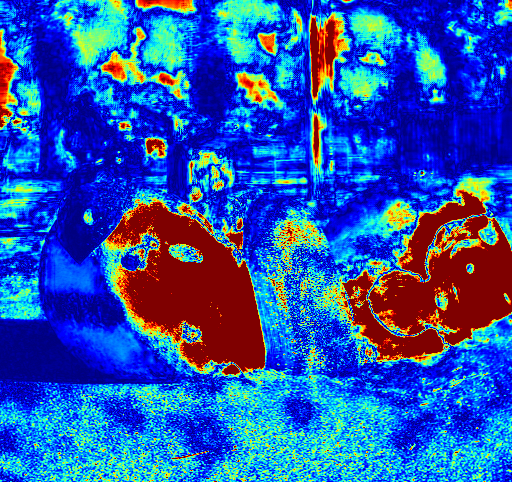}
		\caption{HSCNN+}
		\label{fig:hscnn}
	\end{subfigure}
	\hfill
	\begin{subfigure}[b]{0.14\textwidth}
		\includegraphics[width=\textwidth]{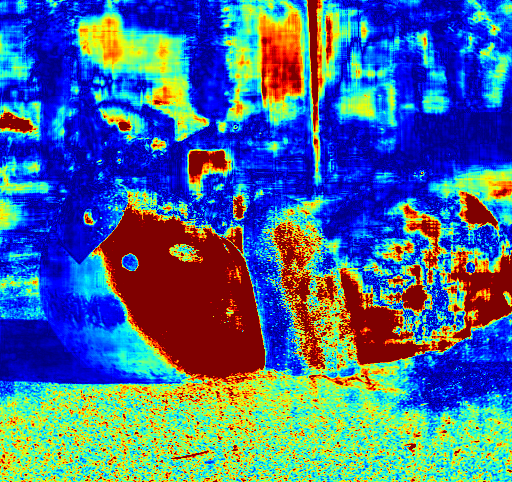}
		\caption{FMNet}
		\label{fig:fmnet}
	\end{subfigure}
	\hfill
	\begin{subfigure}[b]{0.14\textwidth}
		\includegraphics[width=\textwidth]{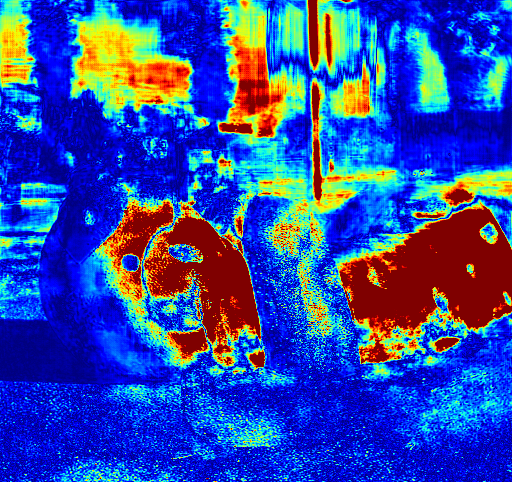}
		\caption{HRNet}
		\label{fig:hrnet}
	\end{subfigure}
	\hfill
	\begin{subfigure}[b]{0.14\textwidth}
		\includegraphics[width=\textwidth]{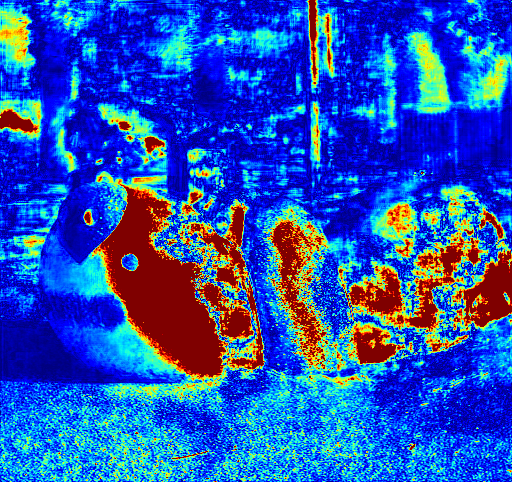}
		\caption{GDNet}
		\label{fig:gdnet}
	\end{subfigure}
	\hfill
	\begin{subfigure}[b]{0.14\textwidth}
		\includegraphics[width=\textwidth]{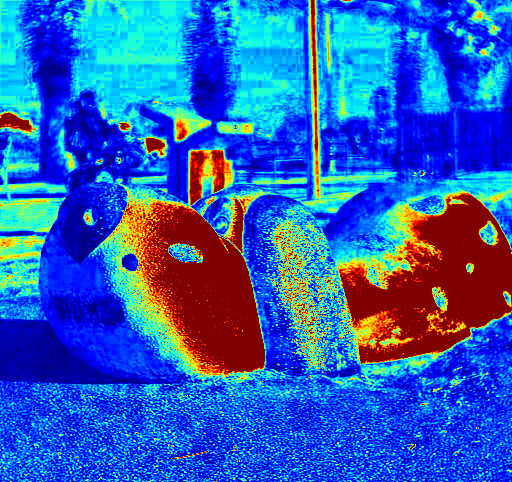}
		\caption{HSRnet}
		\label{fig:hsrnet}
	\end{subfigure}
	\hfill
	\raisebox{0.15\height}{\includegraphics[scale=0.2]{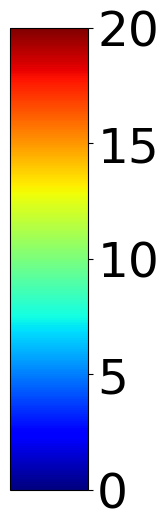}}\\
	
	\begin{subfigure}[b]{0.14\textwidth}
		\includegraphics[width=\textwidth]{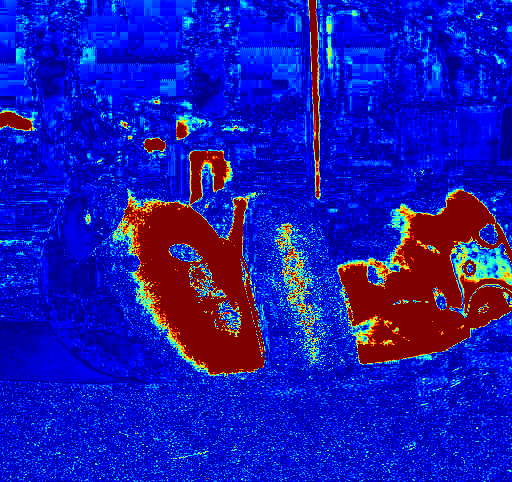}
		\caption{SSDCN}
		\label{fig:ssdcn}
	\end{subfigure}
	\hfill
	\begin{subfigure}[b]{0.14\textwidth}
		\includegraphics[width=\textwidth]{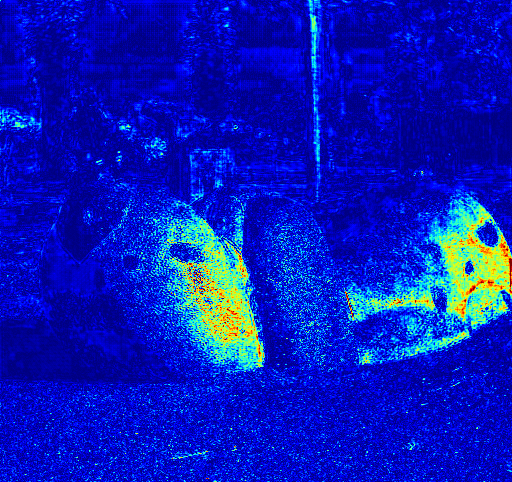}
		\caption{MST++}
		\label{fig:mst++}
	\end{subfigure}
	\hfill
	\begin{subfigure}[b]{0.14\textwidth}
		\includegraphics[width=\textwidth]{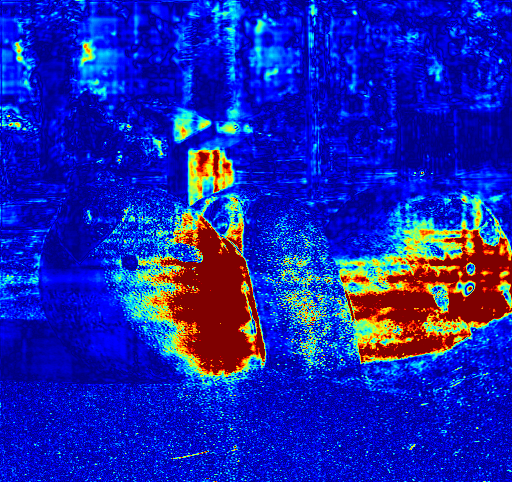}
		\caption{RepCPSI}
		\label{fig:repcpsi}
	\end{subfigure}
	\hfill
	\begin{subfigure}[b]{0.14\textwidth}
		\includegraphics[width=\textwidth]{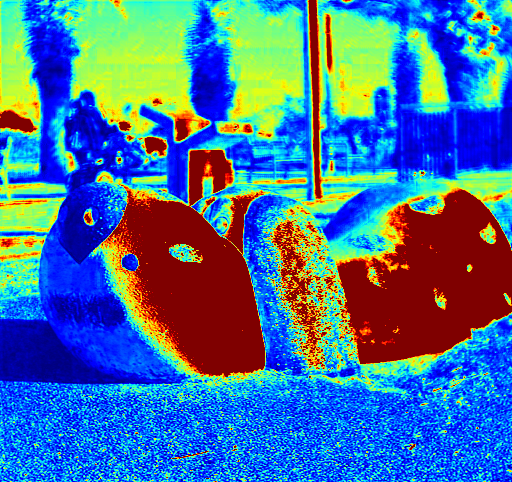}
		\caption{SSRNet}
		\label{fig:ssrnet}
	\end{subfigure}
	\hfill
	\begin{subfigure}[b]{0.14\textwidth}
		\includegraphics[width=\textwidth]{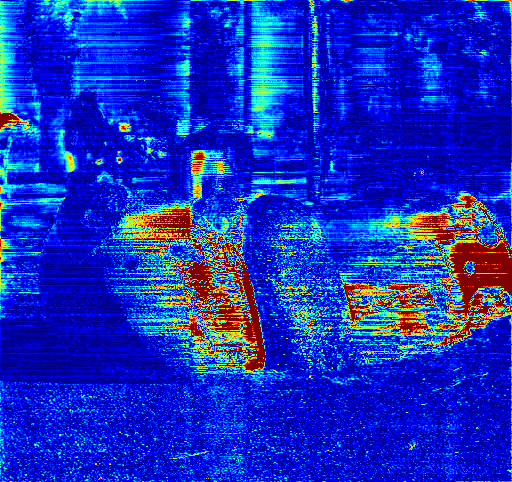}
		\caption{GMSR}
		\label{fig:gmsr}
	\end{subfigure}
	\hfill
	\begin{subfigure}[b]{0.14\textwidth}
		\includegraphics[width=\textwidth]{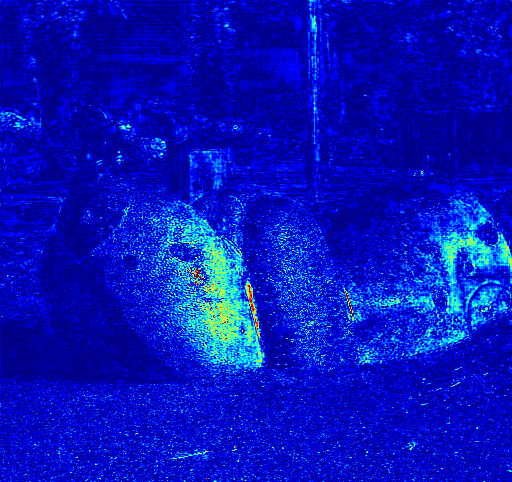}
		\caption{M3SR}
		\label{fig:mssm}
	\end{subfigure}
	\hfill
	\raisebox{0.15\height}{\includegraphics[scale=0.2]{fig/NTIRE2022/ARAD_1K_0907/AE_legend.png}}\\
	
	\caption{Comparison of reconstruction results on the NTIRE2022 dataset using input image ARAD\_1K\_0907 (630~\emph{nm}).}
	\label{fig:heatmaps}
\end{figure*}

\subsection{Dataset and Evaluative Metrics}
To evaluate the performance of M3SR, we conducted extensive experiments using four well-known public SR datasets: NTIRE2022~\cite{Arad2022}, NTIRE2020-Clean, NTIRE2020-Realworld~\cite{Arad2020}, and CAVE~\cite{Yasuma2010}. 
Specifically, the NTIRE2022 dataset contains 900 RGB-HSI training pairs, 50 RGB-HSI validation pairs, and 50 unpaired RGB test images. 
Each of the NTIRE2020-Clean and NTIRE2020-Realworld datasets contains 450 RGB-HSI training pairs, 10 validation pairs and 20 unpaired test images. 
The HSIs in these three datasets have a resolution of \(482 \times 512\) and 31 spectral bands covering 400-700 \emph{nm}. 
The CAVE dataset consists of 32 RGB HSI pairs, each with a resolution of \(512 \times 512\) and 31 spectral bands spanning 400-700 \emph{nm}. 

To ensure fair comparisons, we reorganize the datasets into training, validation and test sets. 
Since the original test images of NTIRE2022 and NTIRE2020 are no longer available after the competitions, we adjust the splits as follows: 
for NTIRE2022, the training set (900 pairs) remains unchanged. 
The validation set of 50 pairs is split into 20 pairs for validation and 30 pairs for testing. 
For NTIRE2020-Clean and NTIRE2020-Realworld, the test set is formed by taking 10 pairs from the validation set and 20 pairs randomly selected from the training set. 
From the remaining training pairs, 30 pairs are randomly selected for validation and the rest remain in the training set. 
For CAVE, we assign three pairs (face, fake\_and\_real\_beers, watercolors) for validation, three pairs (balloons, oil\_painting, fake\_and\_real\_strawberries) for testing, and the remaining pairs for training. 
Following the method in~\cite{Wu2023a}, evaluation metrics include root mean square error (RMSE), peak signal-to-noise ratio (PSNR), spectral angle mapper (SAM), and mean structural similarity (MSSIM). RMSE and PSNR measure numerical accuracy, SAM evaluates spectral fidelity, and MSSIM assesses spatial structure recovery. 
Note that the symbols $\uparrow$ and $\downarrow$ in tables are used to indicate that larger or smaller values of the metric are better, respectively.

\subsection{Implementation Details}
During the training process, RGB images are linearly rescaled to \([0, 1]\), and \(128 \times 128\) RGB and HSI sample pairs are cropped. The batch size is set to 32, and the Adam optimizer with \(\beta_1 = 0.9\) and \(\beta_2 = 0.999\) is used for parameter optimization. The learning rate is initialized to 0.0004 and follows a cosine annealing schedule over 100 epochs. Training data augmentation includes random rotation and flipping. The models are implemented using the PyTorch framework and trained on a single NVIDIA 4090 GPU. In our M3SR implementation, the number of groups is \(G = 4\).

\subsection{Comparisons With SOTA Methods}
To demonstrate the effectiveness of the M3SR, we compare the proposed M3SR with ten state-of-the-art (SOTA) methods on the four benchmarks, including HSCNN+~\cite{Shi2018}, FMNet~\cite{Zhang2020}, HRNet~\cite{Zhao2020}, GDNet~\cite{Zhu2021}, HSRNet~\cite{He2021}, SSDCN~\cite{Chen2021}, MST++~\cite{Cai2022a}, RepCPSI~\cite{Wu2023a}, SSRNet~\cite{Dian2023} and GMSR~\cite{Wang2024a}. 

Table~\ref{tab:NTIRE2022} presents performance comparisons between the proposed M3SR and state-of-the-art (SOTA) SR methods on the NTIRE2022 and CAVE datasets. 
It shows that our method M3SR achieves the best performance in terms of RMSE, PSNR, and MSSIM on the NTIRE2022 dataset, and the best RMSE and PSNR on the CAVE dataset. 
In addition, Table~\ref{tab:NTIRE2022} shows the parameters and computational cost of each method using a 482 $\times$ 512 RGB input, demonstrating that M3SR strikes an excellent balance between computational cost and model performance.
Similarly, Table~\ref{tab:NTIRE2020} shows the performance on the NTIRE2020-Clean and NTIRE2020-Realworld datasets. 
Our method M3SR achieves the best performance in RMSE, PSNR, SAM and MSSIM on the NTIRE2020-Clean dataset and the best RMSE, PSNR and MSSIM on the NTIRE2020-Realworld dataset. These results demonstrate that the proposed M3SR  provides superior reconstruction performance at a significantly lower computational cost.

To further illustrate the superiority of M3SR, we randomly select a spectral band and generated a heatmap of the absolute error between the reconstructed image and the input RGB image for the NTIRE2022 dataset. Fig.~\ref{fig:heatmaps} provides a spatial comparison of the reconstruction performance of the different methods. The bluer the color, the higher the reconstruction accuracy. Obviously, M3SR performs well in reconstructing ARAD\_1K\_0907 (630\emph{nm}) for the NTIRE2022 dataset. 
The observations confirm that M3SR achieves satisfactory reconstruction results, consistent with the quantitative analysis.

\begin{table}[t]
	\centering
	\resizebox{\linewidth}{!}{
		\begin{tabular}{lccccccc}
			\toprule
			Method & SPA & FRE & SPE & RMSE($\downarrow$) & PSNR($\uparrow$) & SAM($\downarrow$) & MSSIM($\uparrow$) \\
			\midrule
			M3SR-V1& & \ding{51} & \ding{51} &0.0381 	&30.4932	&12.5459	&0.8827 \\
			M3SR-V2& \ding{51} &  & \ding{51}  &0.0365	&30.5882	&6.52	&0.9315 \\
			M3SR-V3& \ding{51} & \ding{51} &  &0.0369	&30.3648	&\textbf{\underline{6.2789}}	&0.9241\\
			M3SR & \ding{51} & \ding{51} & \ding{51} &\textbf{\underline{0.0343}}	&\textbf{\underline{31.3995}}	&6.6199	&\textbf{\underline{0.9351}} \\
			\bottomrule
	\end{tabular}}
	\caption{Results of ablation experiments on effects of the spatial perceptual branch (SPA), frequency perceptual branch (FRE), and the spectral perceptual branch (SPE). The best scores are highlighted in \textbf{\underline{underline bold}}.}
	\label{tab:ablation}
\end{table}

\subsection{Ablation Studies}
To validate the effectiveness of the MPF block, ablation studies are provided in this section. Parametric explorations are also provided in order to balance model performance and computational cost. Note that these ablation studies are performed on the NTIRE2022 dataset.

\subsubsection{Effects of Multi-Perceptual Fusion Block} 
This section provides an ablation study to verify the effectiveness of the spatial perceptual, frequency perceptual and spectral perceptual branches. The ablation experimental results of M3SR and its three variants on the NTIRE2022 dataset are presented in Table~\ref{tab:ablation}.

\textbf{Investigation of the Spatial Perceptual Branch:} 
The variant M3SR-V1 is designed by removing the spatial perceptual branch to validate its effectiveness. 
The ablation results show that the absence of the spatial perceptual branch significantly weakens the ability to reconstruct spatial information, as evidenced by a marked worsening of RMSE and PSNR, together with a sharp increase in SAM. 
This indicates that the spatial perceptual branch plays a critical role in the reconstruction of spatial details and structural information.

\textbf{Investigation of the Frequency Perceptual Branch:} 
The variant M3SR-V2 is designed by removing the frequency perceptual branch to validate its effectiveness. 
The ablation results show that the absence of the frequency perceptual branch has a significant impact on the model performance, with RMSE, PSNR, and MSSIM all deteriorating. 
This indicates that the frequency perceptual branch is essential for reconstructing high frequency details and texture information in images.

\textbf{Investigation of the Spectral Perceptual Branch:} 
The variant M3SR-V3 is designed by removing the spectral perceptual branch to validate its effectiveness. 
The ablation results show that the absence of the spectral perceptual branch affects the model performance to some extent, with RMSE, PSNR, and MSSIM all deteriorating. 
This demonstrates that the spectral perceptual branch plays an important role in the reconstruction of spectral information in images.

In summary, these ablation results highlight the importance of all three branches of the MPF block, further validating the effectiveness of the M3SR architecture.

\begin{table}[t]
	\centering
	\resizebox{\linewidth}{!}{
		\begin{tabular}{ccccccc}
			\toprule
			Group & RMSE($\downarrow$) & PSNR($\uparrow$) & SAM($\downarrow$) & MSSIM($\uparrow$) & Params (M) & FLOPs (G) \\
			\midrule
			2 & 0.0351 	&31.2306	&7.7846	 &0.9202	&\textbf{\underline{2.066}}	&\textbf{\underline{91.32}} \\
			4 & \textbf{\underline{0.0343}}	  &\textbf{\underline{31.3995}}	 &\textbf{\underline{6.6199}}   &\textbf{\underline{0.9351}}	&2.166	&100.924 \\
			8 & 0.0372	  &30.6399	 &7.348	   &0.9238	&2.368	&120.133 \\
			16 & 0.0361	  &30.7574	 &11.81	   &0.8793	&2.77	&158.551 \\
			\bottomrule
	\end{tabular}}
	
	\caption{Results of ablation experiments on the number of groups. The best scores are highlighted in \textbf{\underline{underline bold}}.}
	\label{tab:Params}
\end{table}

\subsubsection{Parametric Analysis}
The M3SR relies on a key parameter, the number of groups \( G \) used in the spectral perceptual branch, and parametric analysis is conducted to determine an effective setting for this parameter.  
We evaluate the impact of the number of groups \( G \) on model performance by testing values of 2, 4, 8, and 16. 
As shown in Table~\ref{tab:Params}, the model with \( G=4 \) achieves the best results across most metrics, including RMSE, PSNR, SAM, and MSSIM. 
Although the model with \( G=2 \) is lighter in terms of parameter count and computational complexity, its performance metrics are slightly lower compared to the model with \( G=4 \). 
The models with \( G=8 \) and \( G=16 \) show competitive performance on some metrics, but the number of parameters and computational demands increase considerably as \( G \) grows. 
Therefore, considering both model performance and computational efficiency, setting \( G=4 \) provides a favorable balance and is selected as the preferred configuration.

\section{Conclusion}
This paper presents a multi-scale, multi-perceptual Mamba architecture for SR tasks (M3SR) that addresses the limitations of traditional Mamba. 
The innovative multi-perceptual fusion (MPF) block consists of spatial perceptual, frequency perceptual, and spectral perceptual branches that fully analyze images and capture complex structures. 
By integrating the MPF block into the U-Net structure, M3SR can efficiently capture and fuse global, intermediate, and local features for superior SR. 
Extensive experiments confirm that M3SR not only outperforms existing SOTA SR methods, but also significantly reduces computational cost, making it an efficient solution for SR tasks.

\section{Acknowledgments}
This work was supported by the National Natural Science Foundation of China (NSFC) under Grants 62506238, 62376163, 62173236, 62271324 and 62231020; in part by the Guangdong Regional Joint Foundation Key Project under Grant 2022B1515120076; in part by the Scientific Research Capacity Enhancement Program for Key Construction Disciplines in Guangdong Province under Grant 2024ZDJS063; in part by Shenzhen Technology University School-level Research Project under Grant 20251061020002; and in part by Shenzhen Science and Technology Program under Grant ZDSYS20220527171400002.

\bibliography{SR-2}
\end{document}